\documentclass[conference]{IEEEtran}
\usepackage{cite}
\usepackage{amsmath,amssymb,amsfonts}
\usepackage{algorithmic}
\usepackage{graphicx}
\usepackage{enumitem}
\usepackage{textcomp}
\usepackage{xcolor}
\def\BibTeX{{\rm B\kern-.05em{\sc i\kern-.025em b}\kern-.08em
    T\kern-.1667em\lower.7ex\hbox{E}\kern-.125emX}}
\begin{document}

\title{A Comparative Analysis of CNN-based Deep Learning Models for Landslide Detection}

\makeatletter
\newcommand{\linebreakand}{
  \end{@IEEEauthorhalign}
  \hfill\mbox{}\par
  \mbox{}\hfill\begin{@IEEEauthorhalign}
}
\makeatother

\author{\IEEEauthorblockN{Omkar Oak\IEEEauthorrefmark{1}, Rukmini Nazre\IEEEauthorrefmark{1}, Soham Naigaonkar\IEEEauthorrefmark{1}, Suraj Sawant\IEEEauthorrefmark{1}, Himadri Vaidya \IEEEauthorrefmark{3}}
\IEEEauthorblockA{\IEEEauthorrefmark{1} Department of Computer Science and Engineering, COEP Technological University, Pune, India.}
\IEEEauthorblockA{\IEEEauthorrefmark{3} Department of Computer Science and Engineering, Graphic Era Hill University, Dehradun, India}
}
\maketitle

\begin{abstract}
Landslides inflict substantial societal and economic damage, underscoring their global significance as recurrent and destructive natural disasters. Recent landslides in northern parts of India and Nepal have caused significant disruption, damaging infrastructure and posing threats to local communities. 
Convolutional Neural Networks (CNNs), a type of deep learning technique, have shown remarkable success in image processing. Because of their sophisticated architectures, advanced CNN-based models perform better in landslide detection than conventional algorithms. The purpose of this work is to investigate CNNs' potential in more detail, with an emphasis on comparison of CNN based models for better landslide detection. 
We compared four traditional semantic segmentation models (U-Net, LinkNet, PSPNet, and FPN) and utilized the ResNet50 backbone encoder to implement them. Moreover, we have experimented with the hyperparameters such as learning rates, batch sizes, and regularization techniques to fine-tune the models. We have computed the confusion matrix for each model and used performance metrics including precision, recall and f1-score to evaluate and compare the deep learning models. According to the experimental results, LinkNet gave the best results among the four models having an Accuracy of 97.49\% and a F1-score of 85.7\% (with 84.49\% precision, 87.07\% recall). We have also presented a comprehensive comparison of all pixel-wise confusion matrix results and the time taken to train each model.
\end{abstract}


\section{Introduction}
Landslides pose a significant threat to both human lives and infrastructure, with devastating consequences for communities around the world. Beyond the immediate physical harm, landslides have far-reaching economic and environmental consequences. They disrupt transportation routes, contaminate nearby water sources, and may even trigger secondary hazards such as flooding and tsunamis. \\
According to the WHO, between 1998-2017, landslides have affected an estimated 4.8 million people and have caused more than 18,000 deaths globally. Recovery from landslides involves a comprehensive effort for the restoration of infrastructure as well as the rehabilitation of affected communities. \\
The most common approaches for mapping landslides were visual interpretation of Unmanned Aerial Vehicle(UAV) imagery and conducting field surveys. However, they are restricted because to the inaccessibility of distant locations for field surveys, their dependence on expert experience and knowledge, and their time-consuming, expensive, and inefficient application to broad areas.
In recent years, a large range of Remote Sensing data with varied temporal and spatial resolutions has become accessible due to the significant advancements in Earth Observation technologies. According to Li et al. \cite{1}, because of the extensive side view landslide texture information provided by shipborne images, they may be exploited to obtain high classification accuracy. \\
The implementation of machine learning models for identifying landslides using RS data has mostly been carried out using supervised and unsupervised methods.Unsupervised image classification approaches aggregate pixels with comparable or shared features into the same cluster. K-Means is the most commonly used unsupervised model for landslide mapping. In several research papers, landslides were mapped using unsupervised threshold-based approaches such as change vector analysis, normalized difference vegetation index (NDVI), principal component analysis (PCA), spectral feature variance, and image rationing on multi-temporal pictures. \\
In supervised techniques, the Machine Learning models that have been extensively utilized for modelling and mapping landslides include random forest(RF), decision tree (DT) and support vector machines (SVM). Despite their efficiency in complicated feature identification as well as image classification, these models are prone to problems such as over-fitting, reliance on training data quality, and model setup settings that are inflexible. \\
Deep learning models, such as convolutional neural networks (CNNs), have been applied in a variety of applications during the last decade, notably image processing. Because of the significant recent developments in Remote Sensing technologies and computer vision, the development of powerful graphic processing units(GPUs) and the subsequent availability of large labeled datasets, Deep Learning models have achieved significantly better performance as compared to the conventional ML methods. \\
In this paper, we perform a comparative analysis of four convolutional neural network models - U-Net, LinkNet, PSPNet and FPN for landslide detection through semantic segmentation of satellite images of landslides taken from Bijie, China.

\section{Literature Review}
In the past, various machine learning algorithms were utilized and compared for Landslide detection tasks such as SVM, KNN, random forest, XGBoost, Decision Tree. Faraz S. Tehrani et al. (2021) showed that random forest models can successfully detect landslide segments with a test precision of 96\% \cite{2}.
In a comparison between 6 machine learning algorithms (Logistic Regression, SVM, Random Forest, Gentle Adaboost, LogitBoost, Discrete AdaBoost) and 2 deep learning models (DCNN-11 and CNN-6), DCNN-11 was identified as the most promising model for identifying landslides \cite{3}. \\
Ghorbanzadeh et al. (2019) \cite{4} employed Convolutional Neural Networks for Himalayan landslide detection and compared them to standard machine learning approaches such as support vector machines, random forests and artificial neural networks. The results showed that deep learning models outperform machine learning algorithms and have a superior performance in landslide detection experiments. So, for our comparative analysis, we have preferred Deep learning models based on CNN architecture. Multi layer feed forward neural networks used in CNNs generate accurate image feature perceptions, enabling them to discover visual rules effectively \cite{5}.  \\
In the scope of CNNs, Yang, Shuang, et al. (2022) \cite{6} chose three standard models for semantic segmentation(PSPNet, DeepLabv3+, U-Net) and tested using multiple backbone networks. PSPNet, using ResNet50 as the backbone network, obtained the maximum accuracy with an mIoU of 91.18\%.
FPN was used for unified instance and semantic segmentation on the COCO dataset in 2017 by Kirillov et al. \cite{7}. Additionally, Feature Pyramid Networks(FPNs) \cite{8} have been widely used for semantic segmentation as well. Mask R-CNN, a model for semantic segmentation which was implemented with highly promising results for landslide detection by Ullo et al. \cite{9}, is based on FPN. It has shown high accuracy and precision for landslide detection using various datasets over the past 5 years \cite{10,11},\cite{12}. \\
These deep learning architectures have exhibited substantial success when applied to the analysis of remote sensing imagery. Previous studies have compared different convolutional neural networks (CNNs) for identification of landslides, but there is a lack of comparison specifically focusing on the LinkNet model. It has been used for landslide detection by Ghorbanzadeh et al. in 2022 on the Landslide4Sense dataset along with 10 other state of the art semantic segmentation models \cite{13} and by Garcia et al. in 2023 for the detection of relict landslides. \\
LinkNet showed promising results in both the above implementations, outperforming PSPNet, ContextNet and FCN-8s in terms of recall and F1 score in \cite{13} and FPN in terms of recall in \cite{14}.
A wide variety of landslide datasets have been experimented upon in the past. Images obtained through shipborne photogrammetry in the Three Gorges Reservoir Area in China were used by Yi Li et al. \cite{1} in their paper comparing VGG19, DenseNet121, SEResNet50, Vit and EfficientNetB0. Though the images proved to be adequate for their study, we decided to opt for satellite images due to their high resolution, availability and reliability. \\
Two papers, by Ghorbanzadeh et al.\cite{15} and Meena et al. \cite{16} used images from five optical bands taken by the RapidEye satellite in the Rasuwa district in Nepal to compare the performance of Machine Learning algorithms(RF, SVM) with variants of CNNs. Yang et al.\cite{6} used satellite images from Bijie City, Guizhou Province, China for their study.

\section{Proposed Methods}

\subsection{Data Source}
We used the Bijie landslide dataset, which is an open source satellite imagery dataset, for our research. It includes optical satellite images, shapefiles that depict landslide borders, labels, and digital elevation models.
\begin{figure}[h!]
\centering
\includegraphics[scale=0.04]
{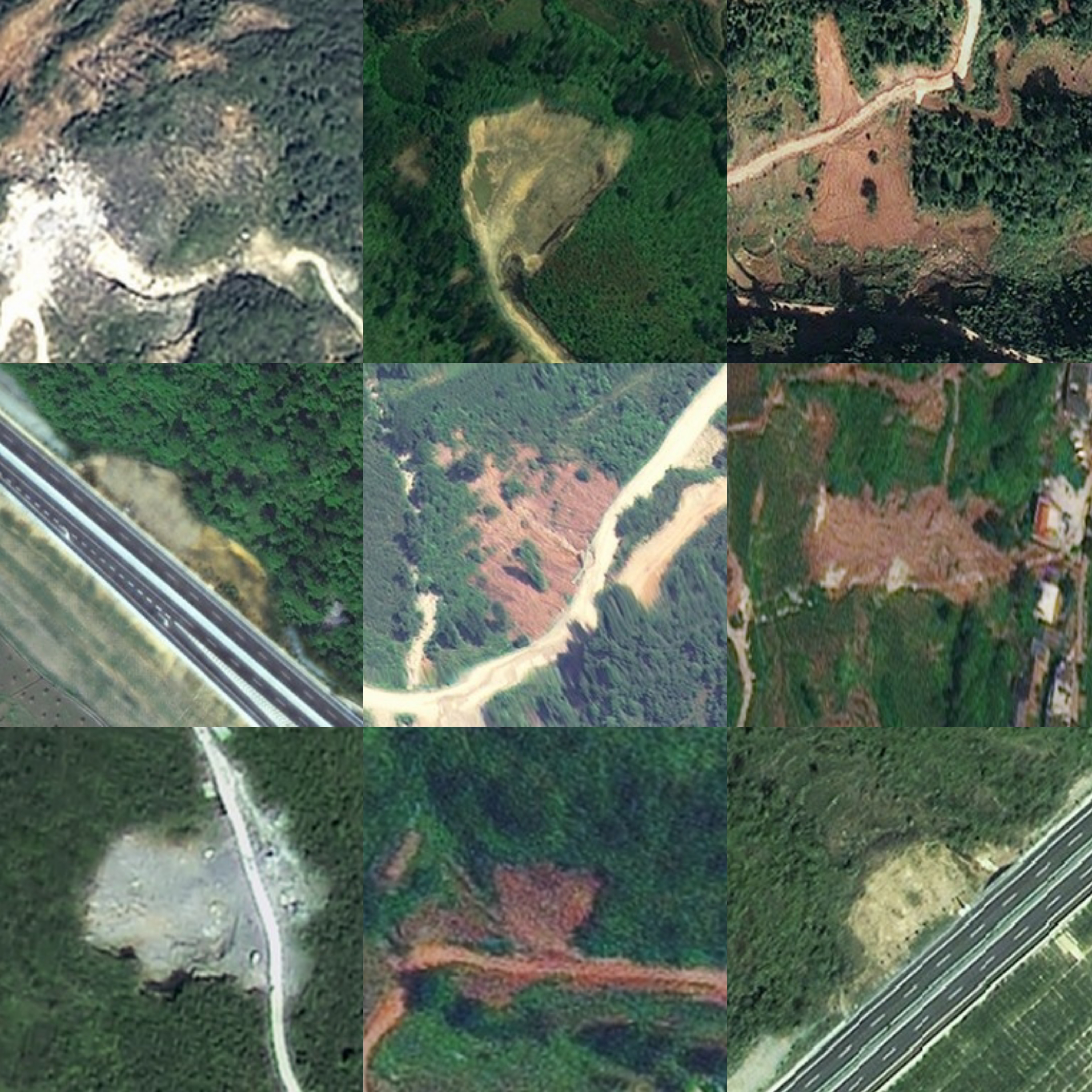}
    \caption{Sample images from Bijie landslide dataset}
    \label{fig1}
\end{figure}
The imagery in this collection were taken by the TripleSat spacecraft between May and August 2018. Its domain of study is in the Bijie City, Guizhou Province, China, at altitudes ranging from 457 m to 2900 m and encompassing around 26,853 square kilometers.  \\
The dataset includes 770 RGB images of landslides and 2003 non-landslide images at a resolution of 0.8 m. Ji et al. (2020) \cite{17} meticulously inspected the data three times before publication. To confirm the database's credibility, two ways of interpreting the landslide photos were used: visual interpretation by geologists using optical remote sensing photographs, and interpretation based on resident accounts and field surveys. ArcGIS was utilized to create mask forms from landslide samples throughout the project. Figure \ref{fig1} shows some sample landslide images from the Bijie landslide dataset.

\subsection{UNet}\label{AA}
Olaf Ronneberger demonstrated U-Net, a semantic segmentation network, at the 2015 ISBI Cell Segmentation Competition \cite{18}. The U-Net approach gathers location and contextual data using a U-shaped network architecture. Initially, it was utilized in the medical industry for image segmentation tasks, especially biological segmentation tasks, and it was eventually applied to other domains including GIS and remote sensing.

\begin{figure}[h!]
\centering
\includegraphics[scale=0.15]
{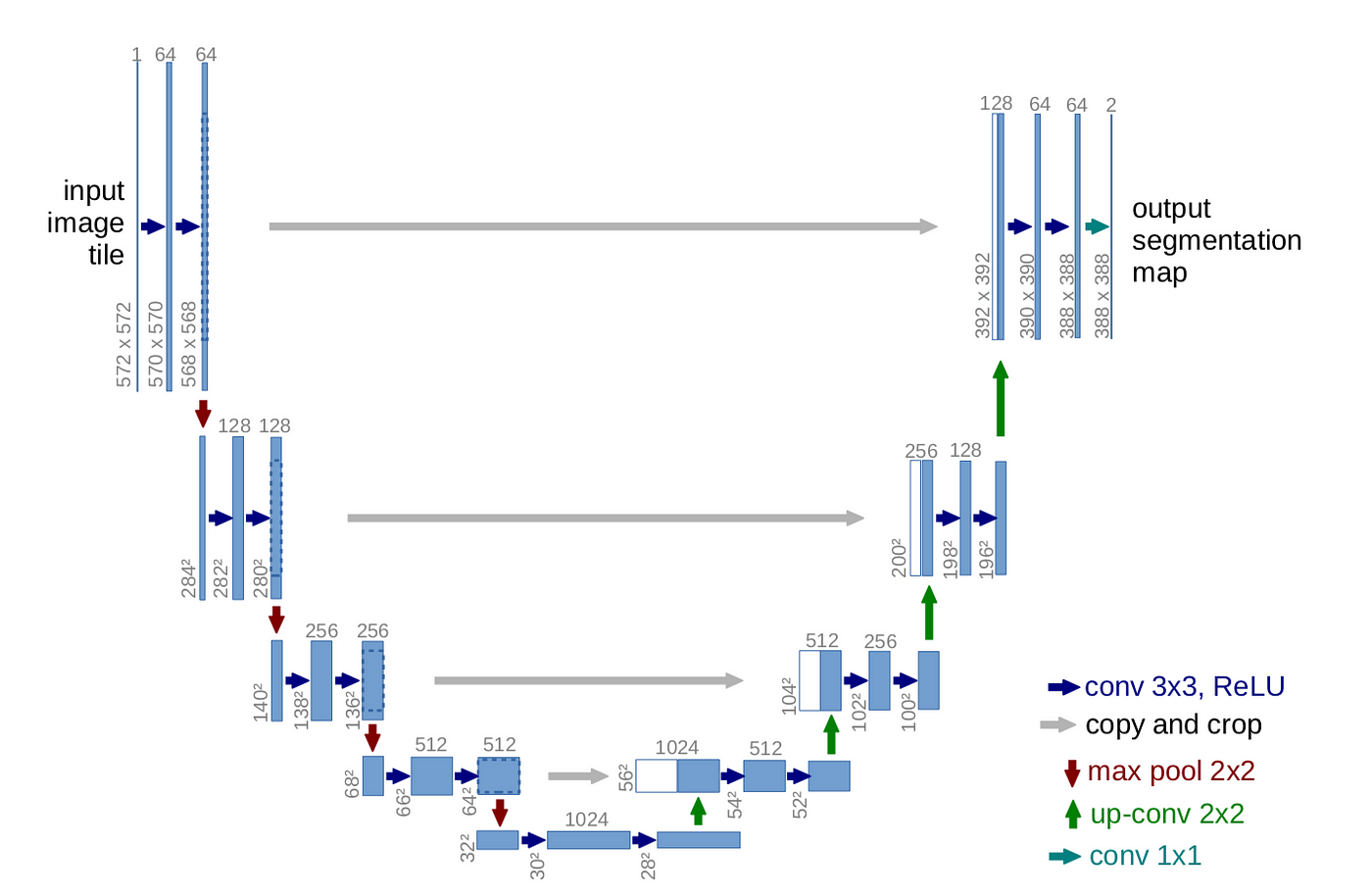}
    \caption{Example of a U-Net architecture}
    \label{fig2}
\end{figure}

Figure \ref{fig2} illustrates U-Net's encoder-decoder network architecture. The encoder uses the concepts of convolutional layer stacking, convolution and pooling to downsample the feature map, and utilizes four pooling operations in total. The size of the feature map is cut in half following each stacking convolution layer operation, and concurrently, the decoder receives the pooling result of each step. Then, in the decoder, the feature map is upsampled and finally concatenated on the channel with the preceding feature map of equal size. The subsequent phase comprises convolution and upsampling, which produces an output image that has the same dimensions as the original image after four rounds of upsampling.

\subsection{LinkNet}
Chaurasia et al. presented LinkNet in 2017 \cite{19} as a semantic segmentation model for real-time interpretation of visual scenes. 
It is a deep neural network architecture that incorporates an efficient information sharing mechanism between the encoder and decoder during each downsampling block as shown in Figure \ref{fig3}. It immediately sends spatial information from the encoder to its equivalent level in the decoder, improving the accuracy of interpretation while reducing the time required for processing. LinkNet can efficiently retain the boundaries of objects in an image, eliminating the need for any additional configurations for training.

\begin{figure}[h!]
\centering
\includegraphics[scale=0.2]
{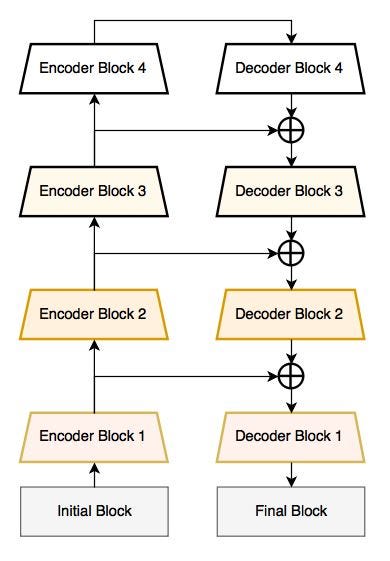}
    \caption{Overview of LinkNet architecture}
    \label{fig3}
\end{figure}

In LinkNet architecture, the initial block features a 7x7 convolution layer with a stride of 2, followed by a 2x2 max-pool layer with a stride of 2. Similarly, the final block conducts a full convolution transitioning from 64 to 32 feature maps, followed by a 2D-convolution. The classifier employs full-convolution with a 2x2 kernel size. \\
For the encoder and decoder blocks, input and output feature map sizes follow the formula $n=64 * 2^i$, where i represents the block index. The first encoder block uses regular convolution without strided convolution, and each convolution layer is followed by batch normalization and ReLU activation, aligning with the ResNet-18 architecture.

\subsection{PSPNet}
Shangtang Technology collaborated with the Chinese University of Hong Kong to create Pyramid Scene Parsing Network (PSPNet), a semantic segmentation model that won the 2016 ImageNet Challenge \cite{20}. PSPNet's first goal was to upgrade the FCN (Fully convolutional network) model. The insertion of a PSP module between the decoder and encoder is the salient feature of PSPNet which also distinguishes it from FCN.  The output of PSPNet is obtained via the convolution process. The framework of the PSPNet model is shown in Figure \ref{fig4} below.

\begin{figure}[h!]
\centering
\includegraphics[scale=0.11]
{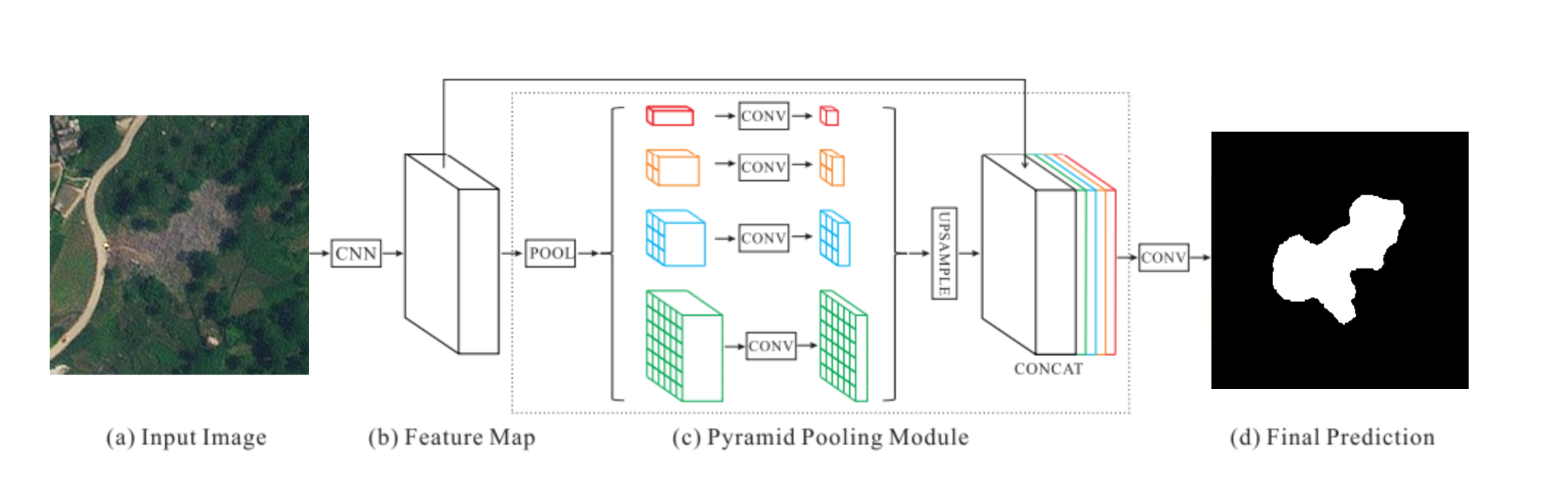}
    \caption{Overview of PSPNet architecture}
    \label{fig4}
\end{figure}

Given an input image, we first use a CNN to generate the final convolutional layer's feature map, followed by a pyramid pooling module to harvest various sub-region representations, which are then combined with concatensation and upsampling layers to form the final representation. The final feature representation includes both global and local context information. The representation is then put through a convolution layer, yielding the final pixelwise prediction.
It should be mentioned that it is possible to customize the size of each level as well as the number of pyramid levels. They are proportional to the size of the feature map taken as input to the pyramid pooling layer. The structure abstracts distinct sub-regions in a few steps utilizing pooling kernels of variable sizes. As a result, the multi-stage kernels should maintain an appropriate representation gap.

\subsection{FPN}
Feature Pyramid Network (FPN) serves as a feature extractor in deep convolutional networks and was first proposed in 2017 by Lin et al. \cite{8} for object detection. It accepts image with any dimension as input and generates feature maps at various layers using fully convolutional algorithms. It works independently of the underlying convolutional architecture, making it a flexible solution that can be integrated into various neural network structures. \\
Figure \ref{fig5} illustrates the pyramid's construction, which includes both a top-down and bottom-up route. The bottom-up approach involves the backbone ConvNet's forward pass. The backbone generates a hierarchy of features using feature maps at various sizes, often using a scaling step of two. Each backbone step has its own pyramid level. 
The output of the last layer of each step serves as a feature map reference set. It begins by upsampling feature maps that are semantically stronger but spatially coarser at higher pyramid levels.
The top-down approach seeks to provide higher-resolution features.

\begin{figure}[h!]
\centering
\includegraphics[scale=0.09]
{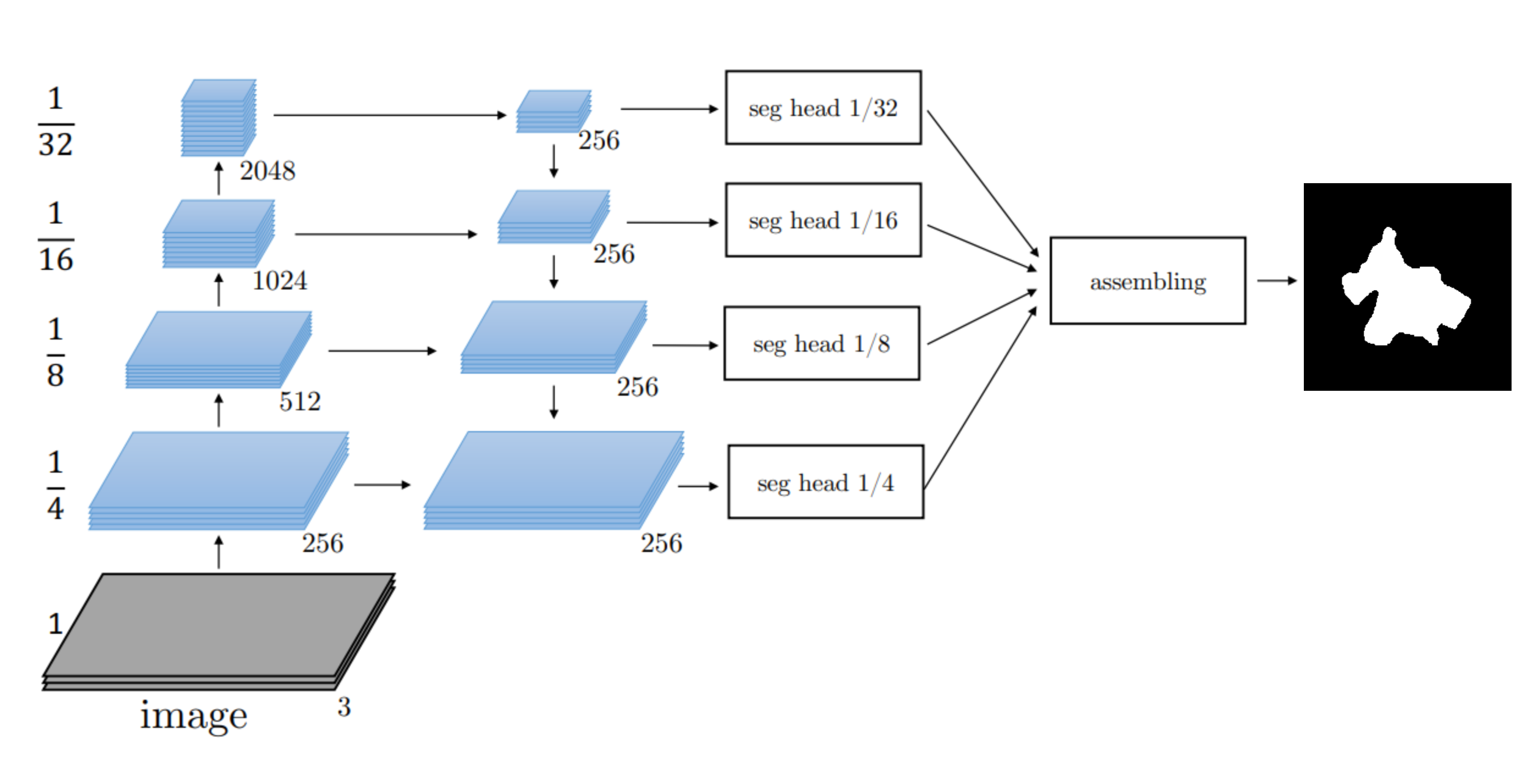}
    \caption{Overview of Feature Pyramid Network (FPN) architecture}
    \label{fig5}
\end{figure}

Then, through lateral connections, these upsampled features are linked with those from the bottom-up pathway. Lateral connections combine feature maps with identical spatial dimensions from both top-down and bottom-up paths. This fusion enhances the feature maps with both high-level semantics from the top-down pathway and accurate localization information from the bottom-up pathway. The final outcome is a feature pyramid with multiple levels, each level containing features that capture information at different scales. FPN's generic nature allows it to be seamlessly integrated with different backbone architectures, providing flexibility in designing neural networks for semantic segmentation tasks.

\subsection{Evaluation Metrics}
In the object detection problem, precision-recall and f-score are the prominent metrics which has been extensively used by the researchers previously \cite{21, 22}, hence considered in this study.

\begin{equation*}
Precision = \frac{TP}{FP+TP} \;\;
Recall = \frac{TP}{FN+TP}
\end{equation*}
Precision is defined as the fraction of real landslide pixels in pixels predicted as landslides by the model, and recall is defined as the ratio of predicted landslide pixels to all actual landslide pixels.

\begin{equation*} 
Accuracy = \frac{TP+TN}{TP+TN+FP+FN}
\end{equation*}

Accuracy is a statistic that describes the model's overall performance across all classes. It is determined as the number of right guesses divided by the total number of predicted outcomes.

\begin{equation*}
F_1=\frac{2}{\text { recall }^{-1}+\text { precision }^{-1}}=2 \frac{\text { precision } * \text { recall }}{\text { precision }+ \text { recall }}
\end{equation*}

The F1-score combines the model's accuracy and recall. The conventional F1-score is calculated by taking the harmonic mean of accuracy and recall. A perfect model has an F-score of one.

\section{Implementation and Results}

\subsection{Preprocessing}
Upon running a trial classification on a few images after resizing, it appeared that the output mask images had more than two unique values. The pixel values were normalized using min-max scaling.The scaled values were then encoded to represent either 0 or 1 for each pixel.

\subsection{Training}
In order to maintain uniformity and ensure reliability of comparison results, the models were implemented in identical environments.\\
The systems employed for the task were equipped with hardware featuring 32GB of RAM and an Intel® Xeon(R) CPU E3-1271 v3 @ 3.60GHz processor. 
In terms of software, the environment consisted of Keras 2.13.1, Tensorflow 2.13.1, and Python 3.8.10.

\begin{table}[h!]
\caption{Model hyperparameter values}
\centering
\begin{tabular}{ll} \hline
\textbf{Hyperparameter}      & \textbf{Parameter Values}      \\ \hline
Input\_shape        & {[}256,256{]}         \\
Classes             & landslide, background \\
Optimizer           & Adam                  \\
Loss                & binary cross entropy  \\
Pretrained weights  & True                  \\
encoder\_weights    & imagenet              \\
Learning rate       & 0.001                 \\
Batch Size          & 16                    \\
Epochs              & 100                   \\
Activation function & sigmoid               \\
encoder\_freeze     & False                 \\ \hline  
\end{tabular}
\label{table2}
\end{table}

In this study, for comparative analysis of our selected models, U-Net, PSPNet, FPN and LinkNet the training parameters were set and adjusted for all the models. ResNet50 was used as the backbone and the input image size was fixed at 256 × 256. Landslide and background were the two output label classes for semantic segmentation. After testing with several learning rates, we fixed the learning rate at 0.001. All the hyperparameter values are kept uniform for all models and are summarized in Table \ref{table2}.

690 images from the Bijie dataset landslide samples were used for training and 80 were used for validation, with an approximate split ratio of 90 -10\% for the dataset containing 770 images. The models were trained for 100 epochs and tested consequently and the trend of training loss for each model was obtained, as given in Figure \ref{fig6}.

\begin{figure}[h!]
\centering
\includegraphics[scale=0.2]
{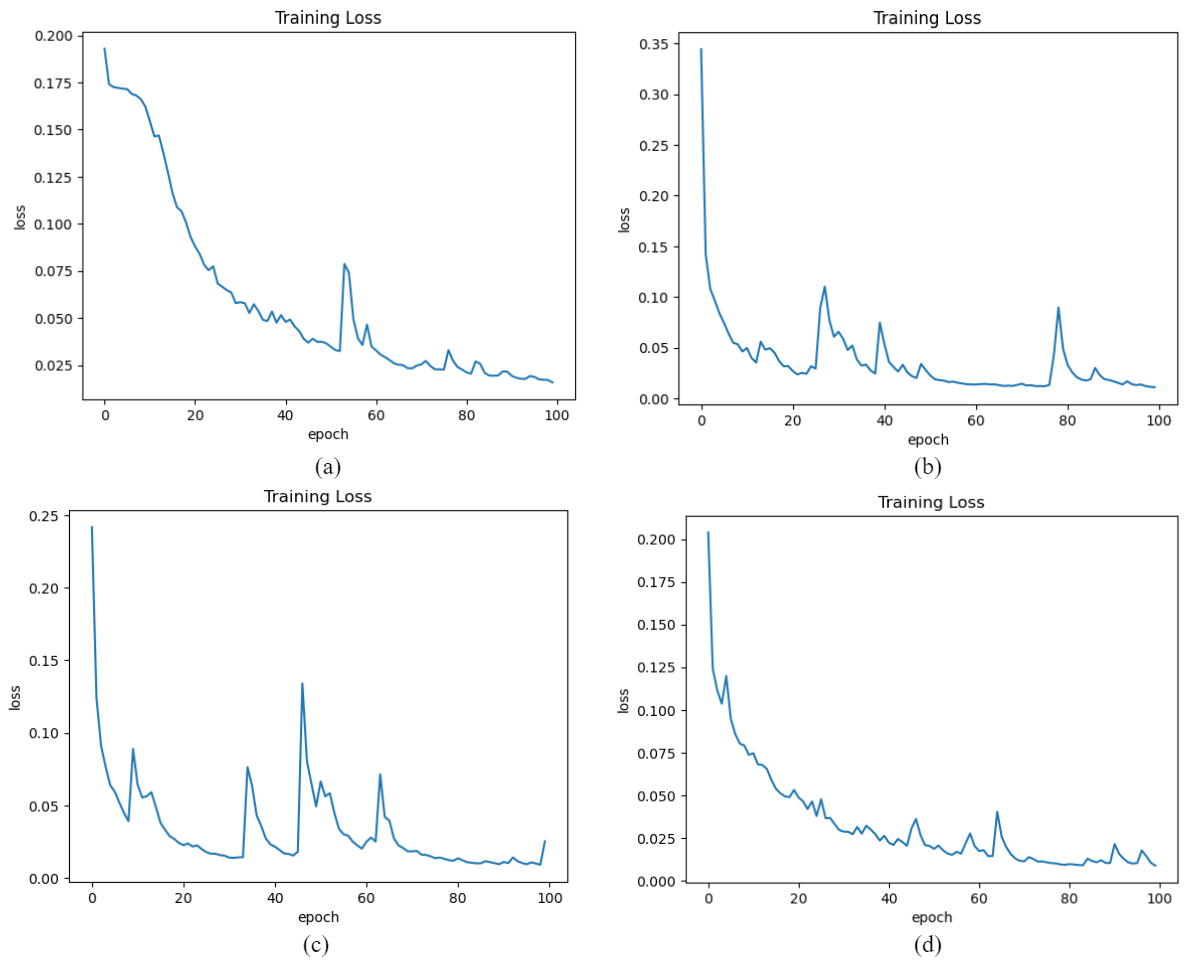}
    \caption{Plots of training loss for the models observed after 100 epochs (a. Unet , b. Linknet, c. PSPNet , d. FPN)}
    \label{fig6}
\end{figure}

All models show promising results with accuracies over 96\% and precision and recall over 80\%. Overall, LinkNet has the best performance with an accuracy of 0.9749 followed closely by Unet with an accuracy of 0.9728. PSPNet and FPN have accuracies of 0.9695 and 0.9689 respectively. Table \ref{table3} contains the performance metrics and the training time(100 epochs) of all four models.

Figure \ref{fig7} shows the confusion matrices obtained for each of the models.

\begin{figure}[h!]
\centering
\includegraphics[scale=0.15]
{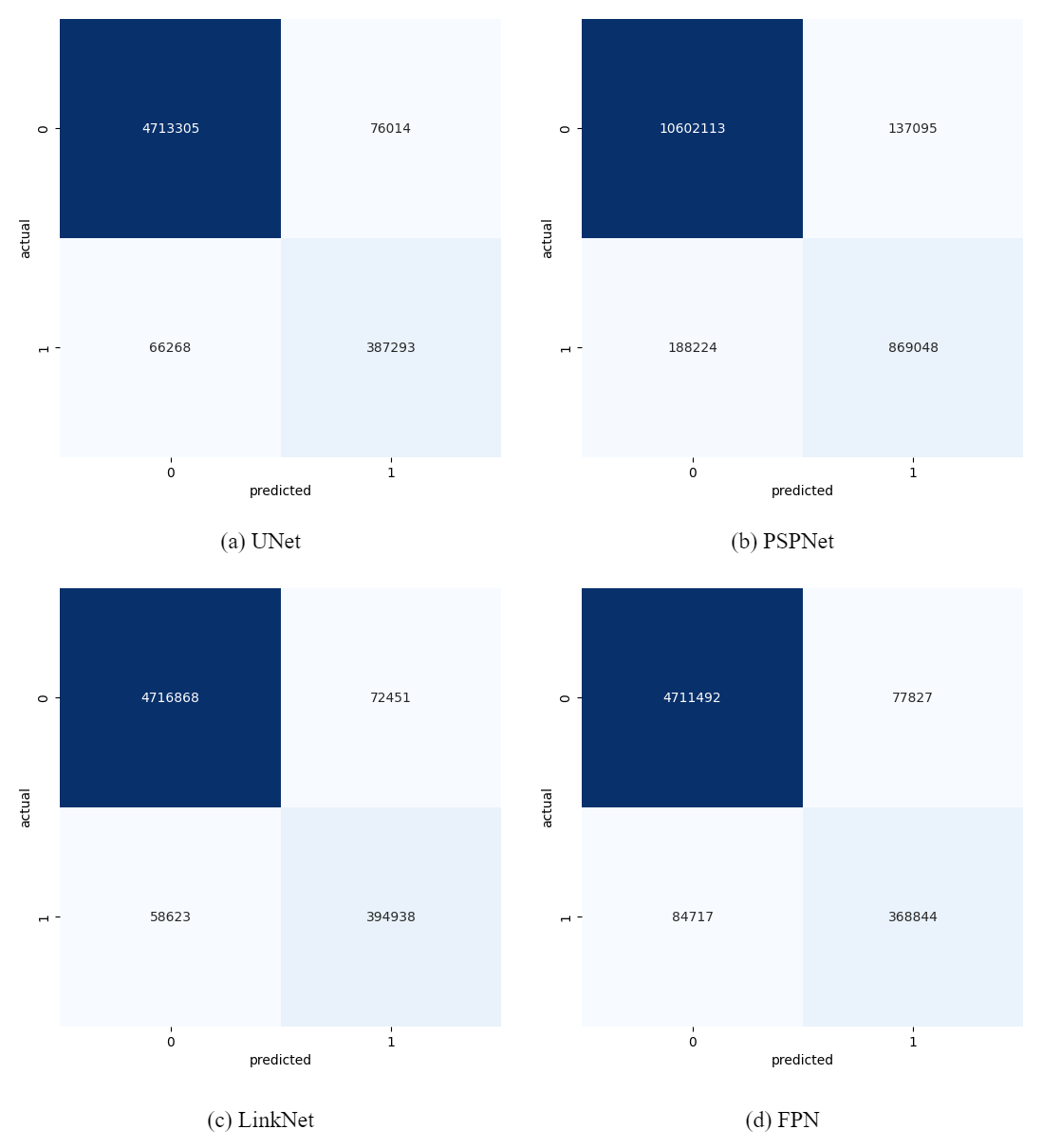}
    \caption{Confusion Matrix for (a) UNet, (b) PSPNet, (c) LinkNet, (d) FPN }
    \label{fig7}
\end{figure}

\begin{table}[h!]
\caption{Comparison of Performance parameters for all models}
\begin{tabular}{lccccc} \hline
Model   & Accuracy & Precision & Recall  & F1 score   & Time        \\ \hline
UNet    & 0.97286  & 0.83593   & 0.85389 & 0.84481    & 661 mins    \\
PSPNet  & 0.96956  & \textbf{0.86374}   & 0.82197 & 0.84234    & 574 mins    \\
FPN     & 0.96899  & 0.82576   & 0.81322 & 0.81944    & 1252 mins   \\ 
LinkNet & \textbf{0.97499}  & 0.84498   & \textbf{0.87074} & \textbf{0.85766}    & \textbf{567 mins}    \\ \hline
\end{tabular}
\label{table3}
\end{table}

All models with the exception of FPN required approximately 10 hours for training. LinkNet required the least time for execution, approximately 574 minutes whereas FPN required more than 20 hours.
UNet and Linknet employ an encoder-decoder architecture (U-shaped) whereas PSPNet and FPN employ a pyramid structured architecture. The U-shaped architectures are designed specifically for semantic segmentation tasks and hence a significant difference is observed between the two types. Moreover, the reason for Linknet performing slightly better than UNet even though their architectures are the same lies in the difference in methods of combining low-level and high-level features.U-Net combines the capabilities of low-level encoders with the features of high-level decoders through skip connections, whereas LinkNet transfers information across the network via link paths.

\begin{figure}[h!]
\centering
\includegraphics[scale=0.1]
{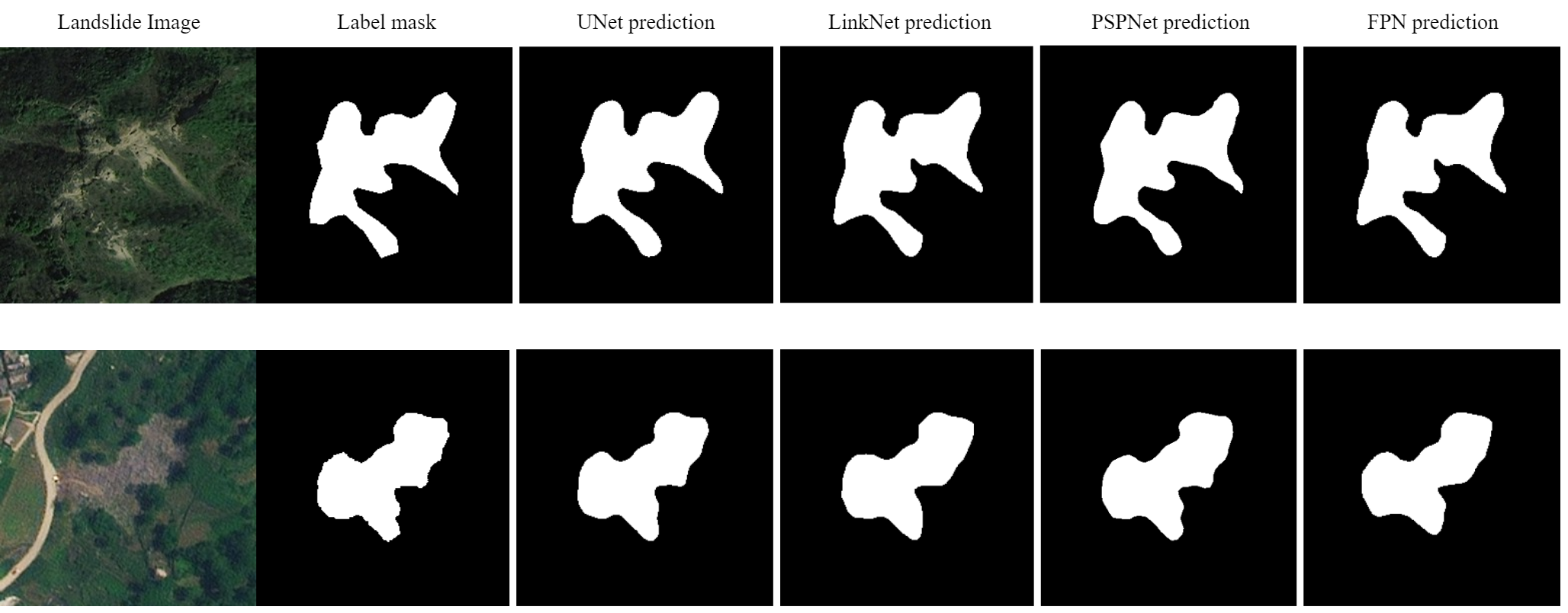}
    \caption{Comparison of label mask with predictions from all models for two images}
    \label{fig8}
\end{figure}

Figure \ref{fig8} shows a comparison between the landslide image, the ground truth and the predicted labels by each model for 2 test images. The labels predicted by LinkNet are slightly better than other models in detecting boundary features.

\section{Conclusion}
In this paper, we offered a comparison of several deep learning-based algorithms for landslide detection and found encouraging results using the Bijie landslide dataset. The dataset was first recreated for semantic segmentation, followed by preprocessing of the landslide data. Then, the models - U-Net, LinkNet, PSPNet and FPN were trained on the Bijie landslide dataset and their performance was compared based on Accuracy, Precision, Recall and F1 score. According to the experimental results, LinkNet achieved the highest accuracy of 97.49\% and F1-score of 85.76\%.
This study shows that it is feasible to use deep learning methods for automatic identification of landslides and that the LinkNet model can reliably identify landslides in real-world scenarios with higher accuracy than other traditional semantic segmentation models. \\
The automatic identification approach may effectively compensate for the drawbacks of manual methods, which are expensive, time-consuming,  and more susceptible to human mistake. Moreover, it can save human resources and time for emergency rescue effort and mitigate losses to property and life. Simultaneously, it can help geological researchers improve the efficiency of their studies and devote more time to field-specific duties. As a result, this discovery has important practical implications. \\
In future work, we plan to extend the applicability of the model by testing its performance in diverse geographical regions with varying terrain, climate, and geological characteristics. This will help validate the generalizability of the model and identify any region-specific adjustments needed. We will further explore the integration of the landslide identification model with other geospatial technologies, such as Geographic Information Systems (GIS) and remote sensing platforms, to enhance the overall understanding of landslide dynamics and improve the decision-making process. \\
We will also attempt to investigate the feasibility of real-time landslide monitoring by integrating the developed model into a broader early warning system. This could involve collaboration with relevant authorities and the implementation of a system that provides timely alerts to mitigate potential disasters.
Our long term objective is to encourage the development of an open-source community around landslide identification research by sharing the codebase, datasets, and methodologies. This will foster collaboration, transparency, and the collective improvement of landslide identification techniques over the years, thereby reducing the damage caused by this formidable natural disaster at a global level.

\end{document}